\documentclass[letterpaper, 10 pt, conference]{ieeeconf}  % Comment this line out if you need a4paper

\IEEEoverridecommandlockouts                              % This command is only needed if 
\usepackage[nolist,nohyperlinks]{acronym}

\usepackage{graphicx}
\usepackage{subfigure}
\usepackage{subcaption}
\usepackage{amsmath}

\usepackage{tabularx}
\usepackage{booktabs}
\usepackage{multirow}
\usepackage{tikz}
\usepackage{amssymb}
\usepackage{pifont}
\usepackage{xcolor}

\usepackage{datetime}
\usepackage{subcaption}
\usepackage{microtype}

\usepackage{url}
\usepackage{hyperref}

\title{\LARGE \bf
\textit{MarineGym}: A High-Performance Reinforcement Learning Platform for Underwater Robotics
}

\author{
Shuguang Chu$^{1,*}$, Zebin Huang$^{2,3,*}$, Yutong Li$^{4}$, 
Mingwei Lin$^{1}$, Dejun Li$^{1}$, \\ 
Ignacio Carlucho$^{2}$, Yvan R. Petillot$^{2}$ and Canjun Yang$^{1}$% <-this % stops a space
\thanks{$^{1}$The State Key Laboratory of Fluid Power and Mechatronic Systems, Zhejiang University, Hangzhou, China}%
\thanks{$^{2}$School of Engineering and Physical Sciences, Heriot-Watt University, Edinburgh, UK}%
\thanks{$^{3}$School of Informatics, University of Edinburgh, Edinburgh, UK}%
\thanks{$^{4}$Department of Mechanical and Aerospace Engineering, the Hong Kong University of Science and Technology, Hong Kong, China}%
\thanks{$^{*}$Co-first authors. Equal contribution}%
\thanks{This research is supported by the Natural Science Foundation of Zhejiang Province, China (Grant LY23E090002); the State Key Laboratory of Deep-sea Manned Vehicles, China (Grant 2024SKLDMV02); and the EPSRC Centre for Doctoral Training in Robotics and Autonomous Systems, United Kingdom (Grant EP/S023208/1).}
\thanks{Mingwei Lin and Dejun Li are corresponding authors.(e-mail:lmw@zju.edu.cn;li\_dejun@zju.edu.cn;)}
}

\begin{document}
\begin{acronym}
    \acro{USV}{unmanned surface vehicle}
    \acro{UAV}{unmanned aerial vehicle}
    \acro{AUVs}{autonomous underwater vehicles}
    \acro{UV}{underwater vehicle}
    \acro{ROV}{remotely operated vehicle}
    % \acro{UUV}{Unmanned Underwater Vehicle}
    \acro{UUVs}{unmanned underwater vehicles}
    \acro{AUV}{autonomous underwater vehicle}
    \acro{ROS}{robot operating system}
    \acro{ROV}{remotely operated vehicle}
    \acro{RL}{reinforcement learning}
    \acro{DRL}{deep reinforcement learning}
    \acro{RAS}{robotics and autonomous systems}
    \acro{RL}{reinforcement learning}
    \acro{USV}{unmanned surface vehicle}
    \acro{UAV}{unmanned aerial vehicle}
    \acro{AUVs}{autonomous underwater vehicles}
    \acro{DRL}{deep reinforcement learning}
    \acro{ROV}{remotely operated vehicle}
    \acro{UUVs}{unmanned underwater vehicles}
    % \acro{UUV}{Unmanned underwater vehicles}
    \acro{NED}{north-east-down}
    \acro{SPH}{smoothed particle hydrodynamic}
    \acro{FEM}{finite element method}
    \acro{DR}{domain randomization}
    \acro{UUV}{unmanned underwater vehicle}
    \acro{PBD}{position-based dynamics}
    \acro{URDF}{unified robot description format}
    \acro{USD}{universal scene description}
    \acro{NN}{neural network}
    \acro{FPS}{frames per second}
    \acro{DDPG}{deep deterministic policy gradient}
    \acro{MDP}{Markov decision process}
    \acro{PPO}{proximal policy optimization}
    \acro{DQN}{deep Q-Network}
    \acro{SAC}{soft actor-critic}
    \acro{TD3}{twin delayed DDPG}
\end{acronym}

\maketitle
\thispagestyle{empty}
\pagestyle{empty}

%%%%%%%%%%%%%%%%%%%%%%%%%%%%%%%%%%%%%%%%%%%%%%%%%%%%%%%%%%%%%%%%%%%%%%%%%%%%%%%%
\begin{abstract}

This work presents the \textit{MarineGym}, a high-performance \ac{RL} platform specifically designed for underwater robotics.
It aims to address the limitations of existing underwater simulation environments in terms of \ac{RL} compatibility, training efficiency, and standardized benchmarking.
\textit{MarineGym} integrates a proposed GPU-accelerated hydrodynamic plugin based on Isaac Sim, achieving a rollout speed of 250,000 frames per second on a single  NVIDIA RTX 3060 GPU.
It also provides five models of \ac{UUVs}, multiple propulsion systems, and a set of predefined tasks covering core underwater control challenges. 
Additionally, the \ac{DR} toolkit allows flexible adjustments of simulation and task parameters during training to improve Sim2Real transfer.
Further benchmark experiments demonstrate that \textit{MarineGym} improves training efficiency over existing platforms and supports robust policy adaptation under various perturbations. 
We expect this platform could drive further advancements in RL research for underwater robotics. 
For more details about \textit{MarineGym} and its applications, please visit our project page: \url{https://marine-gym.com/}.

\end{abstract}

%%%%%%%%%%%%%%%%%%%%%%%%%%%%%%%%%%%%%%%%%%%%%%%%%%%%%%%%%%%%%%%%%%%%%%%%%%%%%%%%
\section{INTRODUCTION}
The vision of deploying \ac{UUVs} capable of performing diverse tasks in unknown and dynamic marine environments is profoundly appealing. Such a robot, capable of independent learning and task execution without human intervention, could revolutionize industries like offshore oil drilling, underwater maintenance, and environmental monitoring by enhancing both safety and efficiency \cite{petillotUnderwaterRobotsRemotely2019a, zengSurveyPathPlanning2015}. 
However, the nonlinear dynamics of \ac{UUVs} and the unpredictable nature of underwater environments pose significant challenges for achieving reliable autonomy. 
These challenges have motivated exploring learning-based approaches, particularly \ac{RL}, which enables robots to learn behaviours through trial-and-error based on predefined reward functions \cite{kober_2013_ReinforcementLearningRobotics}.
Recent RL advancements have enhanced control
tasks for UUVs \cite{petillotUnderwaterRobotsRemotely2019a, walters_2018_OnlineApproximateOptimal, masmitja_2023_DynamicRoboticTracking}.

Despite these advances, applying \ac{RL} in underwater robotics remains challenging due to the complexity of real-world aquatic environments and the risks associated with trial-and-error learning. Consequently, simulation environments offer a controlled and cost-effective platform for developing and testing \ac{RL} algorithms before real-world deployment \cite{liu2021role, ciuccoli2024underwater}. However, existing underwater robotic simulators exhibit significant limitations when interfacing with \ac{RL} frameworks. First, while some simulators, such as HoloOcean \cite{potokar_2022_HoloOceanUnderwaterRobotics} and UNav-Sim \cite{amer_2023_UNavSimVisuallyRealistic}, offer \ac{RL} interfaces, they lack a fully developed RL pipeline and Sim2Real transfer capabilities for real-world deployment. Second, there is a lack of research focusing on baselines and standardized evaluation benchmarks for underwater \ac{RL}, making it difficult to fairly compare different learning algorithms. Third, traditional simulation environments have limitations in large-scale parallel training, which restricts sample efficiency and computational scalability on high-performance GPUs.

\begin{figure}[t]
    \centering
    \includegraphics[width=1.0\columnwidth]{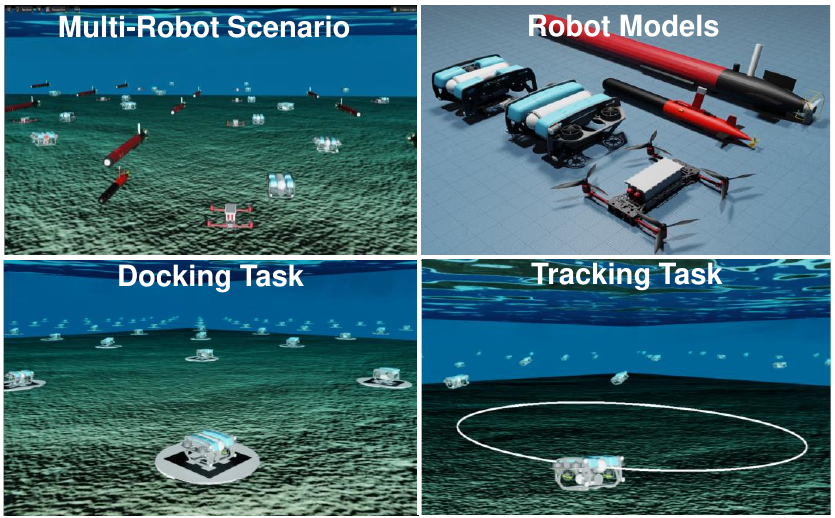}
    \caption{Overview of the \textit{MarineGym} platform featuring underwater robot learning environments. Clockwise from the top left: large-scale multi-robot scenario, underwater robot models, a docking task, and a station-keeping task.}
    \label{Fig:overview}
\end{figure}

To address these challenges, this work introduces \textbf{\textit{MarineGym}}, a specialized platform designed for high-efficiency \ac{RL} training in underwater robotics. \textit{MarineGym} aims to fill the gap in \ac{RL}-oriented training platforms while enhancing existing simulators, rather than replacing them. It enables trained models to be evaluated in existing high-fidelity simulators via Sim2Sim transfer before real-world deployment. In summary, \textit{MarineGym} offers the following key features: 

\begin{itemize}  
    \item Large-scale GPU-accelerated simulation. \textit{MarineGym} employs a GPU-optimized hydrodynamics plugin that supports over 8,000 parallel instances per GPU and generates samples at over 250,000 frames per second. This significantly reduces training time for algorithms such as PPO from hours to minutes.
    \item Flexible training environments. The platform includes five parameterized \ac{UUV} models, four propulsion systems, and standardized tasks. It also provides a \ac{DR} toolkit with tunable parameters for dynamics, sensor noise, and environmental conditions to enhance policy generalization and Sim2Real transfer.  
    \item Standardized benchmarking. \textit{MarineGym} offers a reproducible benchmarking suite with predefined tasks and algorithm baselines. This enables fair performance comparisons for RL research in underwater robotics.
\end{itemize}  

To demonstrate its capabilities and establish baseline performance, we implement and benchmark a range of tasks across different robots, \ac{DR} parameters, and perturbation conditions.

\section{RELATED WORK}

\subsection{Simulated Environments for Underwater Robot Learning} 
After years of development, various underwater robot simulators have emerged. Early simulators, such as UWSim \cite{dhurandher2008uwsim} and UUV Simulator \cite{manhaesUUVSimulatorGazebobased2016}, built on Gazebo-classic, laid the foundation for underwater robotics by providing underwater physics, sensor emulation, and actuator dynamics. However, these simulators lack high-quality rendering and visual fidelity for vision-based applications. Subsequent advancements focused on improving visual realism. Simulators such as URSim \cite{kataraOpenSourceSimulator2019a}, Stonefish \cite{cieslak_stonefish_2019,grimaldi2025stonefish}, and MARUS \cite{loncarMARUSMarineRobotics2022a} integrated modern rendering engines (OpenGL, Unreal Engine 4, Unity3D) to enhance visual fidelity.

Despite advances in physical and visual fidelity, these simulators lack support for robot learning algorithms, particularly \ac{RL}. Most underwater robot simulators are developed within the ROS ecosystem but require additional wrappers to connect to the RL libraries. However, Gazebo-based simulators and other ROS-based architectures face challenges in scalability and sample efficiency. ROS asynchronous execution is not well suited for standard RL implementations and could introduce delays in \ac{RL} interaction \cite{huang2024URoBench}. Chu et al. \cite{chu2024Learning} proposed a parallel ROS-based architecture using DAVE \cite{zhang_dave_2022} to improve performance. Even with parallel execution, the system reached a maximum of 800 \ac{FPS}, which remains insufficient for large-scale RL training. 

In response to these limitations, recent simulators have introduced RL-friendly interfaces to improve accessibility. HoloOcean \cite{potokar_HoloOcean_2022} provides an OpenAI Gym-like API \cite{brockmanOpenAIGym2016a} for RL compatibility and headless execution, removing the need for ROS wrappers. Similarly, UNav-Sim \cite{amerUNavSimVisuallyRealistic2023a} also provides OpenAI Gym support. However, recent work by URoBench \cite{huang2024URoBench} evaluated the popular simulators HoloOcean, DAVE, and Stonefish under RL conditions and identified key limitations. All simulators use CPU-based multiprocessing physics engines, limiting their efficiency in large-scale training data generation. No complete RL pipeline, standardized benchmarks, and baselines have been established. Consequently, RL tasks must be manually defined from scratch, making fair comparisons difficult and hindering reproducibility.

In contrast, GPU-accelerated simulation has significantly improved sample efficiency and scalability in RL-driven robotics research. Platforms such as Isaac Gym \cite{makoviychukIsaacGymHigh2021b} and Orbit \cite{mittal_2023_OrbitUnifiedSimulation} have demonstrated the power of parallelized simulation for aerial robotics \cite{xu_2024_OmniDronesEfficientFlexible, yang2024aam}, humanoid robotics \cite{he_2024_HOVERVersatileNeural} and legged locomotion \cite{hwangboLearningAgileDynamic2019a}. Although these advances tackle computational bottlenecks, existing platforms lack support for underwater robotics, which demands both efficient simulation and a customized RL platform.

\textit{MarineGym} bridges this gap as the first RL platform designed specifically for underwater robotics. It integrates GPU-accelerated large-scale parallel simulation with an RL pipeline, built-in benchmarks, and a DR toolkit for Sim2Real.

\subsection{Reinforcement Learning for Underwater Robots}
The following section reviews RL applications in various \ac{UUV} control tasks, including station-keeping, tracking, and docking. Recent RL advancements have improved performance in these tasks.

\subsubsection{Station-Keeping} RL has been explored for station-keeping to counteract environmental disturbances. Walters et al. \cite{waltersOnlineApproximateOptimal2018} applied model-based dynamic programming, learning a dynamic model in real time and proving stability via Lyapunov theory. Carlucho et al. \cite{carluchoAdaptiveLowlevelControl2018d} extended \ac{DDPG} to full six-degree-of-freedom AUV control, while Knudsen et al. \cite{knudsenDeepLearningStation2019} integrated a dual-controller system where DDPG controlled the horizontal plane and a PD controller stabilized the vertical plane.

\subsubsection{Tracking} Way-point tracking requires reaching targets without predefined paths. Early works discretized the problem using tabular Q-learning \cite{frostEvaluationQlearningSearch2014}, while later studies improved robustness to actuator failures \cite{carluchoAUVPositionTracking2018b}. Model-based methods have also been explored \cite{leonettiOnlineLearningRecover2013, ahmadzadehOnlineDiscoveryAUV2014}, but their reliance on hydrodynamic models limits real-world applicability. Some studies decompose control tasks, such as cable following, and train policies separately using image-based state inputs \cite{el-fakdiTwostepGradientbasedReinforcement2013b}. Other studies leverage policy gradient improvements to enhance stability and convergence speed \cite{shiMultiPseudoQLearningBased2019, yuDeepReinforcementLearning2017}. More recent advancements incorporate nonlinear dynamics \cite{cuiAdaptiveNeuralNetwork2017} and uncertainty-aware control strategies \cite{guoIntegralReinforcementLearningBased2020}, which help reduce oscillations and enhance robustness.

\subsubsection{Docking} This task requires precise motion control to autonomously regulate position, orientation, and velocity for stable alignment with the docking station. Anderlini et al. \cite{anderlini2019Docking} firstly apply \ac{DDPG} and \ac{DQN} to the docking control task. Subsequently, Patil et al. \cite{patil2021Deep} extended it by introducing more algorithms, such as \ac{PPO}, \ac{SAC} and \ac{TD3}. However, they neglect the disturbances of ocean currents. Recently, researchers have proposed improved PPO-based algorithms, such as rollback clipping \cite{zhang2023AUV} and adaptive reward shaping \cite{chu2025Adaptive} to enhance docking success rates and robustness under the influence of currents and waves.

Although RL has shown great potential, its application to UUVs still encounters challenges in real-world transferability, sample efficiency, and safety guarantees. As a benchmarking platform for RL-based underwater control, \textit{MarineGym} provides a set of task environments tailored for underwater robot learning. These environments include tasks such as station-keeping, tracking, and docking designed to evaluate control policies under different hydrodynamic conditions.

\section{METHOD}

\subsection{System architecture}

\begin{figure}[ht!]
    \centering
    \includegraphics[width=1.0\columnwidth]{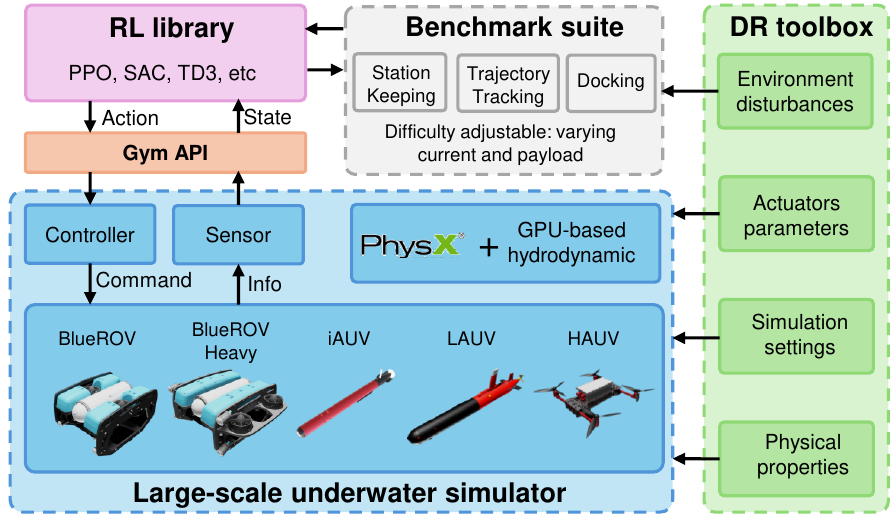}
    \caption{The architecture of \textit{MarineGym}.}
    \label{Fig:architucture}
\end{figure}

The overall architecture of \textit{MarineGym} is shown in Fig. \ref{Fig:architucture}.
It comprises four main components: a GPU-accelerated underwater simulation pipeline, a \ac{DR} toolkit, a multi-type \ac{UUV} model library, and a benchmark suite. The GPU-accelerated underwater simulation pipeline is responsible for creating a hydrodynamic model of \ac{UUV} and the underwater scene. It enables high-fidelity and computationally efficient \ac{UUV} motion simulation by integrating the PhysX \cite{makoviychuk2021Isaac} with a custom GPU-based hydrodynamic plugin. The \ac{DR} toolkit defines various DR parameters, such as physical properties, actuator parameters, simulation settings and environment disturbances. This improves the robustness and Sim2Real transfer ability of the policy. The multi-type \ac{UUV} library includes various types of UUVs, such as ROV-type, torpedo-type, and amphibious models, as well as a variety of underwater actuators. Its modular architecture allows flexible configuration and customization to accommodate different \ac{UUV} designs and research needs.
The benchmark suit consists of several underwater control tasks with varying levels of difficulty, each of which can be configured with ocean currents and payload disturbances to simulate realistic underwater conditions.

\subsection{GPU-accelerated underwater simulation pipeline}

The interaction between fluids and submerged objects remains a challenging problem in computational physics.
Isaac Sim simulates rigid-body dynamics but lacks native support for hydrodynamic forces and moments, which are essential for accurately modelling \ac{UUV} behaviour in underwater environments. 
Various numerical methods have been proposed to address this issue, such as \ac{FEM}. \ac{FEM} divides the computational domain into a finite number of mesh elements, interpolates each element using shape functions, and employs variational methods to solve partial differential equations, like the Navier-Stokes equations. While FEM can achieve high-precision simulations of fluid-object interactions, it relies on iterative solvers and suffers from poor parallel scalability. Another class of methods, known as particle-based approaches, includes \ac{SPH} \cite{monaghan2005Smoothed} and \ac{PBD} \cite{muller2007Position}. These methods track millions of fluid particles, which makes real-time simulation challenging, especially in large-scale multi-robot scenarios. 
To address these limitations, a Newton–Euler equation based on theoretical mechanics analysis has been developed and integrated with Isaac Sim's PhysX engine. Our implementation leverages GPU acceleration for large-scale parallel simulation of thousands of \ac{UUV}. The custom hydrodynamics plugin uses PyTorch tensors for efficient and scalable computation in underwater scenarios. This integration balances accuracy and computational efficiency.

The hydrodynamic calculations of the UUV are based on Fossen's equation of motion \cite{fossen2011handbook}, a well-known theoretical method for UUV simulation. 
The nonlinear equation of motion can be expressed as:

\begin{equation}
\boldsymbol{\tau} = \mathbf{M} \dot{\mathbf{v}} + \mathbf{C}(\mathbf{v}) \mathbf{v} + \mathbf{D}(\mathbf{v}) \mathbf{v} + \mathbf{g}(\boldsymbol{\eta})
\end{equation}

\noindent where $\mathbf{M}$ is the inertia matrix with added mass from surrounding water, $\mathbf{D}$ is the damping matrix, $\mathbf{C}$ is the Coriolis and centripetal matrix, $\mathbf{g}$ is the vector of restoring forces and moments, and $\mathbf{\tau}$ is the actuator's forces.
The \ac{UUV}'s pose is represented by the six-dimensional vector $\boldsymbol{\eta}=[x,y,z,\phi,\theta,\psi]$, where $x,y,z$ denotes its position in the \ac{NED} frame, and $\phi$, $\theta$, $\psi$ correspond to the roll, pitch and yaw angles, respectively.
The body-fixed linear and angular velocity vector is given by $\mathbf{v}$.

To adapt it into the framework, the equation is extended as follows:

\begin{equation}
\begin{aligned}
\mathbf{\tau} = & \underbrace{\mathbf{M}_{RB}\dot{\mathbf{\nu}} + \mathbf{C}_{RB}(\mathbf{\nu})\mathbf{\nu} + \mathbf{g}_{RB}(\mathbf{\eta})}_{\text{Rigid-body term}} \\
& + \underbrace{\mathbf{M}_A\dot{\mathbf{\nu}} + \mathbf{C}_A(\mathbf{\nu})\mathbf{\nu} + \mathbf{D}(\mathbf{\nu})\mathbf{\nu} + \mathbf{g}_A(\mathbf{\eta})}_{\text{Hydrodynamic term}}
\end{aligned}
\label{Equ:fossen}
\end{equation}

The dynamic of the \ac{UUV} in this pipeline is decomposed into two main components: the rigid-body term and the hydrodynamic term.
The rigid-body dynamics are handled by the built-in PhysX physics engine in Isacc Sim, which efficiently computes the UUV's kinematics and collision response.
More importantly, the hydrodynamic effects, which are crucial for modelling underwater motion, are computed using a custom-developed hydrodynamic plugin. This plugin is fully integrated into the simulation pipeline, providing a high-fidelity representation of fluid forces while maintaining computational efficiency. The hydrodynamic forces applied on the \ac{UUV} can be expressed as:

\begin{equation}
\label{Equ:hydra force}
\boldsymbol{\tau}_{hydro} = -\mathbf{M}_A \dot{\boldsymbol{\nu}} - \mathbf{C}_A(\boldsymbol{\nu}) \boldsymbol{\nu} - \mathbf{D}(\boldsymbol{\nu}) \boldsymbol{\nu} - \mathbf{g}_A(\boldsymbol{\eta})
\end{equation}

\noindent The $\mathbf{\tau_{hydro}}$ is calculated and applied to the centroid of each submerged link at each iteration of the simulation pipeline.
It is notable that the $\mathbf{\tau_{hydro}}$ does not include forces and torques from actuators, which will be introduced in Section \ref{Sec:actuators}.
This method of hydrodynamic modelling has been used and validated by other widely adopted underwater simulators, such as the UUV Simulator and UNav-Sim.

Additionally, the ocean currents and mission payloads are provided to simulate the environmental disturbances encountered by \ac{UUVs}. 
Mission payloads are modelled by attaching additional objects to the \ac{UUV}'s body. Their mass and attachment positions are configurable.
The relative velocity of the UUV in the presence of ocean currents is defined as $\mathbf{v_r}=\mathbf{v}-\mathbf{v_c}$.
The term $v_c$ represents the projection of the ocean current velocity in the body-fixed frame.
Consequently, the UUV’s dynamic characteristics are modified by substituting $\mathbf{v}$ with $\mathbf{v_r}$ in the hydrodynamic force equations \ref{Equ:hydra force}.

Constructing a virtual environment that accurately models underwater optical physics is crucial for vision-based underwater tasks.
Due to water's selective absorption and multiple scattering effects on visible light, underwater vision suffers from spectral attenuation and colour distortion.
As illustrated in Fig. \ref{Fig:overview}, \textit{MarineGym} achieves high-fidelity underwater visual effects through an advanced rendering pipeline by leveraging NVIDIA Isaac Sim’s real-time ray-tracing engine.

\subsection{\ac{DR} toolkit}

\ac{DR} is a fundamental technique in \ac{RL} for improving policy robustness and bridging the Sim2Real gap \cite{mehta2020Active}. In \ac{RL}, \ac{DR} improves an agent's generalization ability by introducing diversity into the parameters of the simulation environment. 
The fundamental principle of DR lies in systematically perturbing the parameter space, thereby compelling the policy to learn robust control strategies rather than overfitting to specific environmental configurations.
This necessity becomes particularly pronounced in complex systems such as underwater robotics, where real-world fluid dynamics, mechanical uncertainties, and external disturbances are challenging to model precisely \cite{lu2023Reinforcement}. 
By incorporating randomized training, \ac{DR} effectively covers these long-tail distributions, improving the adaptability and resilience of the learned policies.

To achieve a controllable randomized training environment, the \ac{DR} toolkit is designed based on the following principles:

\begin{itemize}
    \item Modular architecture: The parameter space is divided into four categories: (i) physical properties, including mass, inertia, and centre of gravity; (ii) simulation settings, such as fluid density, added mass matrix, and damping matrix; (iii) actuator parameters, covering time constant, force constant, and installation position; and (iv) external environment, which consists of the velocity and direction of currents, as well as the mass and position of external loads. These parameters can be configured independently or perturbed jointly to introduce variability.  
    \item Flexible sampling strategies: Each parameter supports multiple sampling distributions, including uniform, Gaussian, and custom piecewise functions. The toolkit allows dynamic reconfiguration to adapt sampling distributions.  
    \item Dynamic adjustment of \ac{DR} parameters: Environmental parameters can be modified throughout the training cycle via real-time tuning interfaces. This enables adaptive curriculum learning, where the \ac{DR} range evolves based on policy performance, ensuring a gradual increase in environmental complexity as the agent improves.  
\end{itemize}

\subsection{Multi-type \ac{UUV} model library}
\label{Sec:actuators}

\textit{MarineGym} includes three typical \ac{UUV} models, covering common propulsion configurations and mission scenarios: 

\begin{itemize}
    \item Multirotor type: This type is equipped with multiple thrusters, achieving full-degree-of-freedom control via differential thrust. Two representative models are included: the BlueROV with six thrusters and the BlueROV Heavy with eight thrusters, designed by Blue Robotics \cite{vonbenzon2022OpenSource}. These robots are well-suited for precision operations such as pipeline inspection.
    \item Rudder-propeller type: This configuration combines a forward main thruster with cruciform control rudders, enabling posture control through a coupled thrust-rudder mechanism. Representative models include the LAUV developed by Marine Systems \& Technology Lda \cite{sousa2012LAUV} and iAUV developed by Zhejiang University \cite{chu2025Adaptive}. This type is optimized for high-speed cruising missions.
    \item Tiltrotor type: This type is used in hybrid aerial underwater vehicles with independently tiltable propellers. The HAUV developed by Zhejiang University \cite{zhang2025Adaptive} is included in the platform. It is specifically designed for joint air-sea missions with improved manoeuvrability across aerial and underwater domains.
\end{itemize}

All model parameters are managed in a modular framework with YAML-based customization. Users can adjust thruster type, mass, and inertia, defined in \ac{URDF} files and exported as instanceable assets. This approach enables high-fidelity visualization and supports large-scale simulations.

Underwater actuators commonly include brushless DC motor-driven propellers, rudders, and Servos. All of these types of actuators generate thrust through interactions with the surrounding fluid.
The modelling of the underwater actuator has been extensively studied in the previous literature \cite{manhaesUUVSimulatorGazebobased2016}.
In the \textit{MarineGym}, we decouple their dynamics into two separate modules: the rotor dynamics model, which simulates the time-dependent response of the rotor system, and the thrust generation model, which maps actuator states to the forces and torques.
We consider three actuators with a shared rotor dynamics model, which vary only in specific parameters:

\begin{itemize}
    \item Zero-order model: This idealized static model disregards rotor dynamic response and directly maps input command to steady-state rotational speed.
    \item First-order model: This model utilizes a first-order differential equation to capture the dynamic response, accounting for inertial and lag effects. It offers a representation of motor acceleration and deceleration.
    \item \Ac{NN} date-driven model: This model utilizes a \ac{NN} to learn the nonlinear mapping between input commands and rotor dynamics from experimental data. It captures complex effects such as friction hysteresis and damping. Instead of explicit differential equations, the \ac{NN} provides a black-box function approximation.
\end{itemize}
The thrust generation model of propellers is represented as a quadratic function with a dead zone, where the thrust is proportional to the deviation of rotational speed beyond predefined thresholds, accounting for nonlinear efficiency variations.
The thrust generation model of rudders follows a simplified lift and drag formulation, where the generated forces depend on fluid velocity, rudder surface area, and lift and drag coefficients, which vary with the angle of attack.
All actuator parameters are fully configurable, which allows researchers to adapt flexibly to different actuator models. 
To facilitate practical implementation, we provide built-in parameters for two widely used thrusters, T200 and 2820 from Blue Robotics, from experimental data collected using an underwater thrust testing platform.

In the \textit{MarineGym} platform, the type, number, and installation positions of actuators are predefined in the configuration file. 
During the simulation, each actuator is assigned a unique index. At each simulation step, the force $\tau_i$ generated by each actuator is first computed based on the input commands and the actuator model. 
Then, the computed forces are individually applied to the corresponding actuator frame.
The decoupled design enhances modularity for independent actuator refinement, while flexible parameterization enables seamless adaptation to various vehicles and missions with minimal reconfiguration.

\section{Experiments}

\subsection{Simulation speed comparative}

To evaluate the efficiency and practicality of \textit{MarineGym}, we conduct experiments with the BlueROV2 Heavy model on the station-keeping task.
A key aspect of this evaluation involved comparing the \ac{FPS} achieved by \textit{MarineGym} against several widely used UUV simulation platforms.
The hardware setup in our experiments includes a 16-core AMD Ryzen 7950X processor, an NVIDIA RTX 3060 GPU, and 12GB of memory. 
The system operates on Ubuntu 20.04 as the underlying operating system. 
For robotics simulation and control, we utilize the Robot Operating System (ROS) Noetic and Gazebo 11.0 as the simulation environment.

As shown in Table \ref{Tab:simulation Speed comparation}, it presents a comparative analysis of simulation speed across different platforms. Due to variations in test conditions across DAVE, Stonefish, and HoloOcean, we focus on comparing the order of magnitude rather than exact values. \textit{MarineGym} supports over 8000 environments on a single GPU, achieving up to 250,000 \ac{FPS}. However, the FPS does not scale linearly with the number of environments due to GPU performance constraints. The upper limit of environments was not tested, as our focus was not on pushing the maximum scale. Increasing the number of environments introduces additional overhead from communication and scheduling. This results in a trade-off between the total number of environments and the FPS per environment. 

In contrast, existing simulation platforms such as Stonefish, DAVE, and HoloOcean have not been optimized for parallel acceleration due to their inherent architectures \cite{huang2024URoBench}.
Previous efforts to improve DAVE’s performance using CPU-based multithreading and TCP/IP parallelization allow it to support up to 32 parallel environments, achieving a peak speed of approximately 1,000 FPS \cite{chu2024Learning}. However, due to system communication bandwidth constraints, further scaling remains limited.  These comparisons underscore \textit{MarineGym}’s superior efficiency in high-speed, large-scale \ac{RL} training for underwater robotics.

\begin{table}
\centering
\normalsize
\caption{Comparison of simulation speed (order of magnitude). $\mathcal{O}(10^n)$ denotes an approximate scale.}
\begin{tabular}{ccc} 
\toprule
\textbf{Platform} & \hfil \textbf{Env num} & \hfil \textbf{Speed (FPS)} \\
\midrule
\multirow{3}{*}{\textit{MarineGym}} & $2.0 \times 10^3$ & $\sim 1.7 \times 10^5$ \\
 & $4.1 \times 10^3$ & $\sim 2.1 \times 10^5$ \\
 & $8.2 \times 10^3$ & $\sim 2.5 \times 10^5$ \\
Parallelized DAVE & $\mathcal{O}(10^1)$ & $\mathcal{O}(10^3)$ \\
DAVE & $\mathcal{O}(10^0)$ & $\mathcal{O}(10^1)$ \\
Stonefish & $\mathcal{O}(10^1)$ & $\mathcal{O}(10^2)$ \\
HoloOcean & $\mathcal{O}(10^1)$ & $\mathcal{O}(10^2)$ \\
\bottomrule
\end{tabular}
\label{Tab:simulation Speed comparation}
\end{table}

\subsection{Benchmarking on UUV tasks}

Based on the previously introduced platform, we establish a benchmark comprising three control tasks of varying difficulty levels on five different types of \ac{UUV}s. All tasks are formulated as \ac{MDP} and are accessible to \ac{RL}algorithms via a Gym-style interface. Each task follows an end-to-end design principle, where the state space includes only the UUV's intrinsic kinematic information along with essential task-related parameters. The action space comprises control commands for all actuators, which establish a direct mapping from state observations to control actions. The specific tasks are described as follows:

\begin{itemize}
    \item Station-keeping: As a fundamental underwater control task, this task requires the \ac{UUV} to approach and maintain a target pose in a dynamic ocean environment. This task demands both position error convergence and compliance with attitude constraints, primarily evaluating the system's resistance to flow disturbances.
    \item Trajectory tracking: This task requires the \ac{UUV} to follow predefined time-varying three-dimensional trajectories (e.g., helical curves, Lissajous curves). It focuses on assessing the agent's capability to predict and track time-varying reference signals.
    \item Docking: This task is designed for underwater docking, which requires the \ac{UUV} to land precisely on an underwater platform while maintaining stability upon contact. The challenges include precise position adjustment, attitude control, and robustness against flow disturbances.
\end{itemize}

\begin{table}
\centering
\normalsize
\caption{Environment Parameter Settings. Parameters marked with (*) represent relative ratios to their original values, while unmarked parameters represent absolute values.}
\begin{tabular}{cccc} 
\toprule
\textbf{Parameters} & \shortstack{Train Env\\ Range} & \shortstack{Test Env1\\ Range} & \shortstack{Test Env2\\ Range} \\
\midrule
Model mass*     & [0.8, 1.2] & 1.1 & 1.4 \\
Volume*         & [0.8, 1.2] & 1.1 & 1.4 \\
CoBM            & [0.5, 3.0]  & 2.0 & 4.0 \\
Inertia*        & [0.8, 1.2] & 1.1 & 1.4 \\
Added mass*     & [0.8, 1.2] & 1.1 & 1.4 \\
Damping*        & [0.8, 1.2] & 1.1 & 1.4 \\
Current velocity  & [0.0, 0.5]  & 0.2 & 0.8 \\
Payload mass*   & [0.0, 0.3] & 0.2 & 0.4 \\
\bottomrule
\end{tabular}
\label{Tab:env_parameters}
\end{table}

Experiments are conducted on five distinct types of UUVs mentioned above to systematically evaluate the performance of different UUV models across the three benchmark tasks.
Each task is configured with three levels: (i) standard environment, (ii) environment with disturbances such as ocean currents and payload variations, and (iii) environment with disturbances and \ac{DR}.
The parameter ranges for disturbance, and \ac{DR} are listed in Table \ref{Tab:env_parameters}.
These configurations are designed to assess the robustness of \ac{RL} agents under progressively challenging conditions. For each task-environment combination, experiments were conducted with five different random seeds to account for variability in training outcomes.

\begin{figure*}[ht!]
    \centering
    \includegraphics[width=1.0\textwidth]{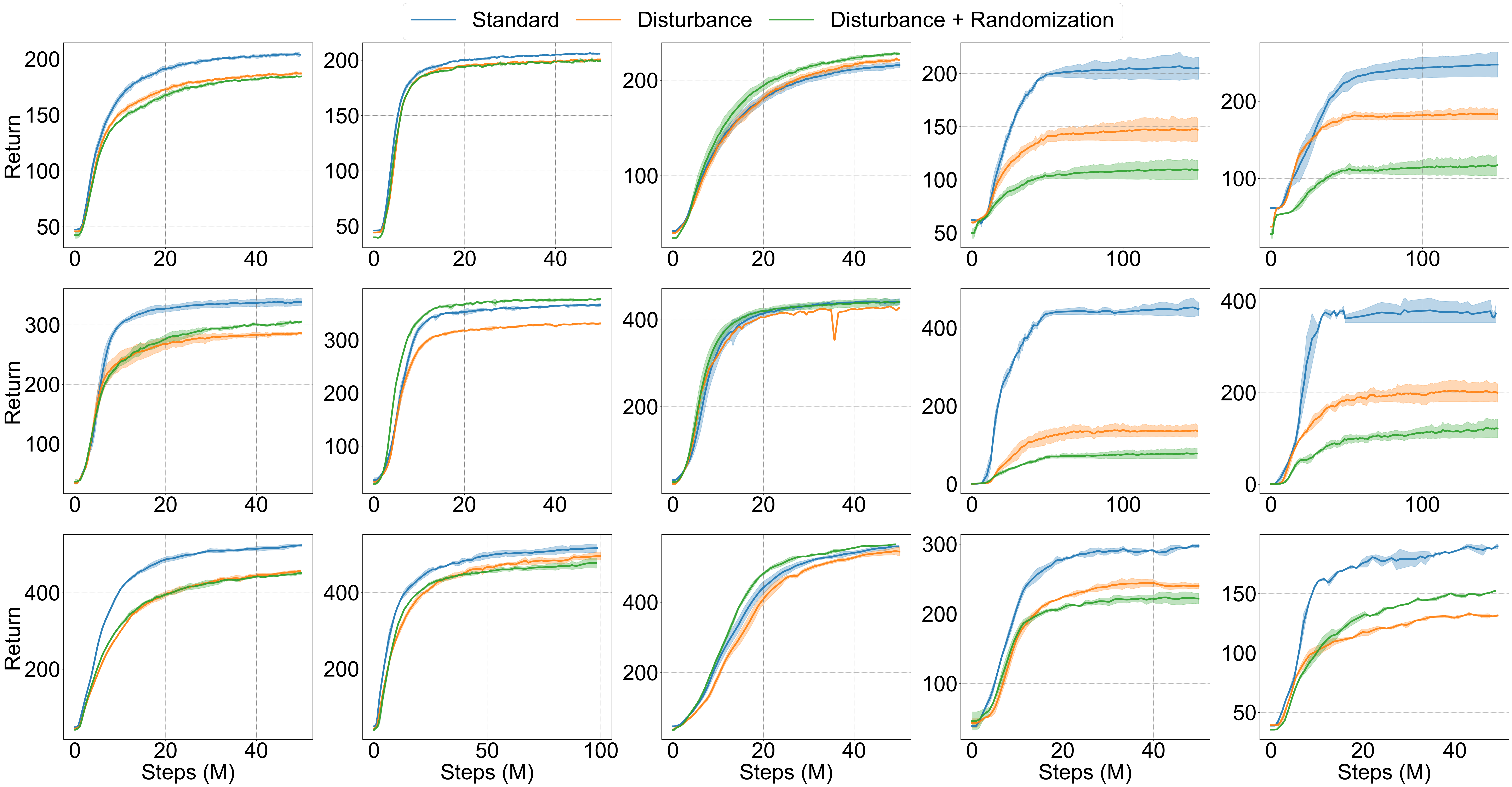}
    \caption{Learning curves of five UUV models across three tasks under Standard (blue), Disturbance (orange), and Disturbance + Randomization (green) environments, each trained with four different random seeds. Each row represents a different task (Station keeping, Trajectory tracking, and Docking), while each column corresponds to a specific UUV model (BlueROV, BlueROV Heavy, HAUV, LAUV, iAUV). }
    \label{Fig:benchmark}
\end{figure*}

The learning curves across all experiments are presented in Fig. \ref{Fig:benchmark}.
All of them demonstrate convergence to maximum return values during the later training phases, with minimal standard deviation indicating consistent performance across multiple random seeds.
However, LAUV and iAUV require approximately twice the training steps of other \ac{UUVs}, primarily due to their underactuated dynamics.
Additionally, a general performance degradation across most experimental scenarios when environmental disturbances and \ac{DR} are introduced. 
The BlueROV Heavy and HAUV platforms, both equipped with eight-actuator configurations, exhibit minimal performance deterioration, which can be attributed to their enhanced manoeuvrability characteristics. 
The BlueROV, with six actuators, shows slightly lower robustness.
The LAUV and iAUV perform significantly worse, struggling to maintain position and stability under strong ocean currents.

The benchmark results show that all UUVs converge to optimal performance. However, underactuated models (LAUV and iAUV) require significantly longer training, reflecting the control challenges in their dynamics. Performance degradation is observed when disturbances and \ac{DR} are introduced, while heavily actuated vehicles (BlueROV Heavy and HAUV) exhibit greater robustness due to improved manoeuvrability.

\subsection{\ac{DR} evaluation}

To assess the DR toolkit’s impact on policy robustness and adaptability, we conduct a Sim2Sim evaluation across the three tasks: station keeping, trajectory tracking, and docking. The evaluation follows a structured protocol:

\begin{itemize}
    \item For each task, we train two policies: one with DR and one with NDR (No Domain Randomization).
    \item Both policies are deployed in two distinct environments: Env1, which falls within the DR parameter range, and Env2, which extends beyond the DR range. The parameter settings are detailed in Table \ref{Tab:env_parameters}.
    \item Each policy is evaluated through 500 independent trials, and the average performance is quantified using task-specific error metrics: (i): Station keeping error: The Euclidean distance between the UUV and the target position. (ii) Trajectory tracking error: The mean deviation across all tracking points along the predefined trajectory. (iii) Docking error: The Euclidean distance between the UUV and the centre of the docking platform upon contact.
\end{itemize}

As shown in Table \ref{Tab:dr_evaluation}, policies trained with DR exhibit greater robustness across varying simulation environments compared to those with NDR. 
In all three tasks, DR-trained policies achieve significantly lower error values in Env1, which lies within the DR training range. This demonstrates a strong ability to adapt to known environmental variations. 
For Env2, which represents conditions outside the DR range, both policies experience some degradation in performance. 
However, DR-trained policies still outperform the NDR counterparts, maintaining lower error values across all tasks. 
Notably, for the docking task, DR reduces error by up to 64\% compared to NDR in Env2, highlighting its effectiveness in improving policy adaptability to novel conditions.

\begin{table}
\centering
\normalsize
\caption{Comparison of policy performance with DR and NDR (No Domain Randomization)}
\begin{tabular}{lccc} 
\toprule
\textbf{Task} & \textbf{Settings} & \hfil \textbf{Error (Mean $\pm$ Std)} \\
\midrule
\multirow{4}{*}{\shortstack{Station\\Keeping}} 
 & NDR (Env1) & $1.6370 \pm 0.7916$ \\
 & NDR (Env2) & $1.8980 \pm 0.8364$ \\
 & DR (Env1) & $\mathbf{0.0361 \pm 0.0290}$ \\
 & DR (Env2) & $0.1466 \pm 0.2439$ \\
\midrule
\multirow{4}{*}{\shortstack{Trajectory\\tracking}} 
 & NDR (Env1) & $0.0929 \pm 0.1030$ \\
 & NDR (Env2) & $0.1017 \pm 0.1165$ \\
 & DR (Env1) & $\mathbf{0.0474 \pm 0.0554}$ \\
 & DR (Env2) & $0.0989 \pm 0.1081$ \\
\midrule
\multirow{4}{*}{Docking}  
 & NDR (Env1) & $0.0885 \pm 0.0438$ \\
 & NDR (Env2) & $0.2410 \pm 0.2080$ \\
 & DR (Env1) & $\mathbf{0.0401 \pm 0.0157}$ \\
 & DR (Env2) & $0.0876 \pm 0.1245$ \\
\bottomrule
\end{tabular}
\label{Tab:dr_evaluation}
\end{table}

\section{Limitation and future work}

The \textit{MarineGym} platform is proposed in this research to optimize \ac{RL} training for underwater robotics. \textit{MarineGym} features large-scale GPU-based parallel simulation, a native Gym-compatible interface, and an integrated \ac{DR} toolkit. Extensive evaluations across multiple models and tasks demonstrate the platform’s high efficiency and scalability in RL for \ac{UUV}s.
Additionally, Sim2Sim evaluation shows that \ac{DR} significantly enhances policy robustness and generalization, allowing RL agents to adapt more effectively to diverse and perturbed underwater conditions.
However, the platform lacks validation in real-world environments. 
Moreover, the platform supports only basic task types, and key underwater visual characteristics, such as light propagation and turbidity, are not fully replicated.
We hope that \textit{MarineGym} will serve as a foundation to inspire researchers to further explore RL applications in underwater robotics, ultimately enhancing its performance.
Future work will focus on Sim2Real validation and the development of complex, realistic underwater scenarios to extend applicability to more challenging tasks.

\bibliographystyle{ieeetr}

\clearpage

\end{document}